\documentclass[a4paper]{article}

\usepackage[english]{babel}
\usepackage[utf8x]{inputenc}
\usepackage[T1]{fontenc}

\usepackage[a4paper,top=3cm,bottom=2cm,left=3cm,right=3cm,marginparwidth=1.75cm]{geometry}

\usepackage{amsmath}
\usepackage{graphicx}
\usepackage[colorinlistoftodos]{todonotes}
\usepackage[colorlinks=true, allcolors=blue]{hyperref}
\usepackage{times}  % DO NOT CHANGE THIS
\usepackage{helvet} % DO NOT CHANGE THIS
\usepackage{courier}  % DO NOT CHANGE THIS
\usepackage[]{graphicx} % DO NOT CHANGE THIS
\usepackage{graphicx}  % DO NOT CHANGE THIS
\usepackage{amsfonts}
\usepackage{booktabs}
\usepackage{amsmath}
\usepackage{subcaption}

\usepackage[square]{natbib}

\title{Text classification with pixel embedding}

\author{\Large \textbf{Bin Liu\textsuperscript{\rm 1,2}, Guosheng Yin\textsuperscript{\rm 2}, Wenbin Du\textsuperscript{\rm 2}}\\
\textsuperscript{\rm 1} Center of Statistical Research,
School of Statistics\\
Southwestern University of Finance and Economics\\
\textsuperscript{\rm 2} Department of Statistics and Actuarial Science\\
The University of Hong Kong\\
}

\begin{document}
\maketitle

\begin{abstract}
We propose a novel framework to understand the text by converting sentences or articles into video-like 3-dimensional tensors. Each frame, corresponding to 
a slice of the tensor, is a word image that is rendered by the word's shape. The length of 
the tensor equals to the number of words in the sentence or article. 
The proposed transformation from the text to a 3-dimensional tensor makes it very convenient to implement an $n$-gram model with convolutional neural networks for text analysis. Concretely, we impose a 3-dimensional convolutional kernel on the 3-dimensional text tensor. The first two dimensions of the convolutional kernel size equal the size of the word image and the last dimension of the kernel size is $n$. That is, every time when we slide the 3-dimensional kernel over a word sequence, the convolution covers $n$ word images and outputs a scalar. By iterating this process continuously for each $n$-gram along with the sentence or article with multiple kernels, we obtain a 2-dimensional feature map.
A subsequent 1-dimensional max-over-time pooling is applied to this feature map, and three fully connected layers are used for conducting text classification finally.
Experiments of several text classification datasets demonstrate surprisingly 
superior performances using the proposed model in comparison with existing methods.
\end{abstract}

\section{Introduction}
Word representation, is the foundation of natural language processing (NLP) tasks, such as the text classification \citep{kim2014convolutional}, machine translation \citep{sutskever2014sequence}, question answering \citep{zhou2015simple}, etc.
The most straightforward approach to word representation is the one-hot encoding, which projects words into sparse 1-of-$V$ vectors with $V$ being the size of the vocabulary.
Another popular framework of word representation is to construct word vectors using the word2vec \citep{mikolov2013distributed,pennington2014glove}, which is an unsupervised approach. Both the 1-of-$V$ and word2vec encodings have their own limitations. For example, the one-hot embedding has the issue of curse-of-dimensionality and word2vec requires the availability of a prior corpus for pre-training. Both one-hot and word2vec encodings are word-level embeddings.
In addition to the word-level encoding, \citet{zhang2015character} propose a character-level convolutional neural network (char-CNN) which quantifies the characters for alphabetic scripts. However, the character-level embedding is inapplicable to ideograph languages such as Chinese and Japanese, because the number of characters for such languages can be huge.

Intuitively, when we read an article on a screen, our eyes capture the text as a series of images which are then passed onto the brain for recognition and understanding. 
Hence, a natural way of word representation is to use visual shapes of the words or characters 
as features \citep{shimada2016document,sun2019vcwe,su2017learning,liu2017learning}.
For examples, \citet*{su2017learning} and \citet*{shimada2016document} take Chinese and Japanese characters as images and apply a subsequent convolutional autoencoder to take those images as inputs and then output low-dimensional character embeddings. With such character embeddings, a char-CNN \citep{shimada2016document} or traditional recurrent neural networks  \citep{su2017learning,liu2017learning} can be adapted for Chinese and Japanese text analysis tasks.

\begin{figure}[htp!]
    \centering
    \includegraphics[width=9cm]{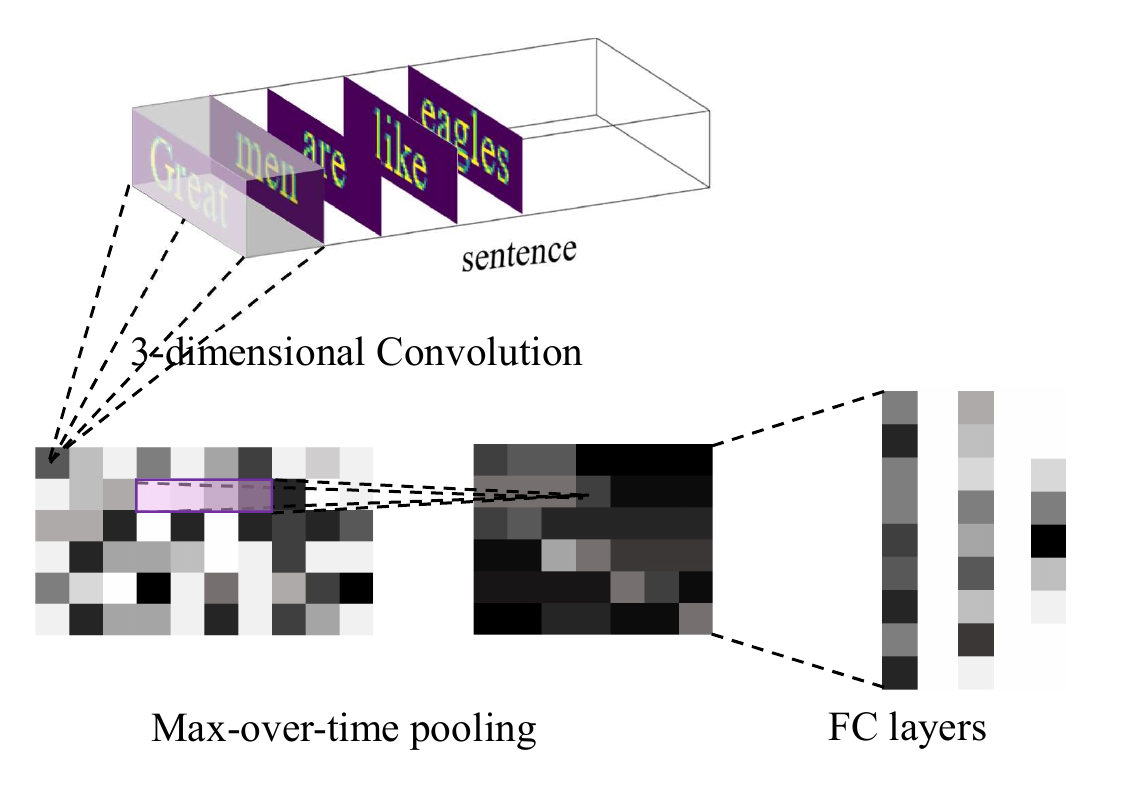}
    \caption{An illustration of a 2-gram model with max-over-time pooling. Words in a sentence or
    an article are sequentially stacked into a video-like tensor. A 3-dimensional convolutional kernel operates on two neighboring word images at a time which acts as a 2-gram detector. The convolutional kernel covers the whole word in each slide, which is different from the small 2-dimensional kernel in \citep{liu2017learning}. The max-over-time pooling is then applied to each row of the feature matrix separately, and finally three fully connected layers unfold
    the feature map as inputs and output the classification results.}
    \label{fig:model}
\end{figure}

However, the limitations of these visual embedding models are obvious: 
(1) Characters are treated separately with a traditional local convolutional 
kernel, which ignores the statistics of characters or possibilities of words' co-occurrence  
($n$-gram characteristics); (2) These models compress the word or character's visual 
vector into a low-dimensional vector, which makes these models lack interpretability;
and (3) Existing visual embedding based models are all designed for ideograph languages. 
To solve these problems, we propose a novel framework to adapt 
a word's pixel (visual) embedding for English, while 
our model can be easily extended to any other languages.
Concretely, we render the shape of a word $v_i$ in a document or a sentence $S$ as an image $X_i \in \mathbb{R}^{w\times h}$ and then fold those images into a 3-dimensional tensor sequentially. That is, the document or sentence is converted into a video-like 3-dimensional tensor. Each frame of the video corresponds to a word and the length of the video equals to the number of words in the document. To capture the $n$-gram characteristics of the text, we propose to impose 3-dimensional convolutions on the ``text video". Compared with a small convolutional kernel (traditionally, a $3\times3$ kernel is most frequently used), we use a big 3-dimensional kernel of size $w\times h \times n$ to extract 
statistics information of the text, where $w$ and $h$ are the width and height of word images
respectively, and $n$ is the number of words covered by the kernel. 
This can be interpreted as an $n$-gram model as shown in Figure \ref{fig:model}. With multiple 3-dimensional kernels, the convolutional layer outputs a feature matrix, whose columns are features of $n$-grams and rows correspond to the channels of different kernels. Following \citep{kim2014convolutional}, a subsequent 1-dimensional max-over-time pooling is applied to this 2-dimensional feature map. Finally, three fully connected (FC) layers are used for conducting text classification.

The contributions of our work are three-fold:
\begin{enumerate}
     \item We propose to represent a sentence or an article with a video-like 3-dimensional tensor, and each frame of this tensor represents one word in the sentence or article;
    \item We use a 3-dimensional convolutional kernel to learn the $n$-gram features from the tensor representation of the text;
    \item We evaluate our model on several text classification tasks on both performances and interpretability.

    % robust pixel representation for characters, which not only makes the text preprocessing unnecessary but also enhances the model interpretability;
    % \item (2) we design a simple convolutional neural network (CNN) to accept the text's pixel embedding as input, and this model demonstrates surprisingly excellent results on Chinese text classification.
\end{enumerate}
\section{Related Works}
Recently, deep learning has been shown to achieve impressive performances on the NLP
tasks \citep{kim2014convolutional,wang2015semantic,iyyer2015deep,goldberg2016primer,jiang2018text,jacovi2018understanding,tang2019entity}. Under the NLP framework via deep neural networks, one typically needs to find a way to embed the raw text into features that computers can ``recognize and understand''. Currently, the existing approaches for text embedding can be categorized into three frameworks from coarse to fine. 
The first one is the document-level or sentence-level approach 
that embeds documents or sentences into vectors \citep{le2014distributed,lin2017a}.
The second one is the word-level embedding \citep{mikolov2013distributed,pennington2014glove,joulin2017bag}, and the last one is character-level \citep{zhang2015character} or radical-level embedding \citep{ke2017radical}.

The simplest implementation of the word-level embedding is to encode words as one-hot vectors. The dimension of the one-hot vector equals the size of the vocabulary $V$. Typically, $V$ ranges from thousands to tens of thousands, which may hence lead to the issue of curse-of-dimensionality. Another approach is to construct a corpus-related matrix that contains statistical information of this corpus and then compute
the word representation by factoring the matrix \citep{deerwester1990indexing}.
However, the size of the constructed matrix is usually large, which makes the decomposition very time-consuming. The most classical way of word-level embedding is based on the word2vec framework \citep{mikolov2013distributed}, which is
originated from the neural language model. The word2vec encodes semantic features of words into a low-dimensional dense vector with the word's local context, which is also called distributed word vectors.  However, the quality of word vectors heavily depends on the quality and quantity of the corpus. As an improvement, \citet{pennington2014glove} propose to incorporate the global matrix factorization \citep{deerwester1990indexing} and local context \citep{mikolov2013distributed}, which can strike a balance between the performance and cost. For an NLP task, both the one-hot and distributed representation methods have their own limitations: the one-hot embedding has the issue of curse-of-dimensionality and the word2vec requires the availability of a corpus as well as pre-training prior to a specific NLP task.

Another popular framework is the document to vector or sentence to vector \citep{le2014distributed,lin2017a}, which aims to represent sentences, paragraphs, and documents with vectors. \citet{le2014distributed} propose a ``Paragraph Vector" model to learn fixed-length feature representations for sentences or documents in an unsupervised way. The ``Paragraph Vector" is based on word2vec \citep{mikolov2013distributed}.

\citet{zhang2015character} propose a character-level encoding model that quantifies the characters in English words sequentially. Combining this elegant design of text embedding with convolutional neural networks (CNN), their method achieves excellent results on text classification. Unfortunately, the character-level encoding method is only applicable to the phonogram, such as English, but cannot be extended to logogram languages, such as Chinese or Japanese. Following the char-CNN, \citet{ke2017radical} propose to encode Chinese and Japanese characters with the semantic radical components to bridge this gap. In their model, each Chinese or Japanese character can be divided into a sequence of radical-level embeddings. However, the radical-level method ignores the spatial structure of Chinese characters, which is a big difference between Chinese and alphabetic scripts. The pixel embedding proposed in this paper is completely
different from the existing encoding methods. It is directly motivated by the way of human reading, for which eyes receive visual signals of the text and then send them to the brain for further analysis. Therefore, we use the pixel image of the text as its representation 
%following 
which exactly mimics the way how human read the text.

%Different from 
In contrast to the one-hot, word2vec, and char-level embedding, 
human read and understand the text from a completely different perspective,
which is based on the visual shapes of the words.
Intuitively, when we read a web article on the screen or a book, our eyes capture the text as a series of images rather than embedding them into vectors. In other words, human understand the text with the visual information of the words, i.e., we recognize characters or words from their images that are captured by our eyes. Therefore, we believe that the pixel image, i.e., the character's morphological shape, provides a natural way to represent characters and words. Motivated by this idea, several visual embedding methods \citep{shimada2016document,su2017learning,sun2019vcwe} have been developed for 
Chinese and Japanese text understanding.
However, it is very difficult to visually embed alphabetic languages such as English, 
because English words cannot be rendered as the same sized image as Chinese or Japanese
characters.

\begin{table*}[!ht]
  \caption{Comparisons among the one-hot vector (1-of-$V$), character-level CNN (char-CNN \citep{zhang2015character}), word2vec (distributed representation, \citep{mikolov2013distributed}), radical-level method (radical-level, \citep{ke2017radical}), and our method
  in terms of preprocessing steps for word vectors. Note that $\surd$ indicates
  the procedure that must be taken;
  %or has that characteristic;
  -- indicates not applicable; blank means not needed.}
%do not have to carry out.}
  \label{tb:com-preprocess}
  \centering
  \begin{tabular}{lccccc}
    \toprule
    % \multicolumn{}{c}{Part}                   \\
    % \cmidrule(r){1-2}
   & Our method   & Char-CNN     & 1-of-$V$ & Distributed & Radical-level \\
    \midrule
    Slangs and abbreviations &   &      & $\surd$ & $\surd$ &     \\
    \midrule
    Remove noise &    &      & $\surd$ & $\surd$ & $\surd$    \\
    \midrule
    Remove stop words &    &      & $\surd$ & $\surd$ & $\surd$    \\
    \midrule
    Remove low-frequency words &    &      & $\surd$ & $\surd$ &     \\
    \midrule
    Stem and lemmatization & --   &      & $\surd$ & $\surd$ & --    \\
    \midrule
    Remove misspellings &    &      & $\surd$ & $\surd$ & $\surd$    \\
    \midrule
    Pre-training &    &      &  & $\surd$ &     \\
    \midrule
    Maintain a vocabulary &    & $\surd$     & $\surd$ & $\surd$ & $\surd$    \\
    \midrule
    Sparsity of vector & $\surd$   & $\surd$     & $\surd$ &  & $\surd$   \\
    \midrule
    Dimension of vector  & $10^3$   & 70     & $10^3 \sim 10^4$ & $10^2$ & $10^3$    \\
    \bottomrule
  \end{tabular}
\end{table*}

\section{Model}
\subsection{Overview}

The proposed model for the text classification is shown in Figure \ref{fig:model}. Given a document or a sentence $S$, we first render the word $v_i$ in this document as a matrix $X_i \in \mathbb{R}^{w\times h}$. Sequentially, a series of text matrices $\{X_1, X_2, \ldots, X_l\}$ are then folded into a 3-dimensional tensor $\mathcal{X} \in \mathbb{R}^{w\times h \times l }$, where 
$l$ is the length of the sentence. 
In other words, the document or a sentence $S$ is taken as a ``video", and each frame of the video corresponds a word of $S$ with the size of $w \times h$.
Compared with extracting a word's representation from the visual pixel map with a convolutional autoencoder \citep{shimada2016document,su2017learning}, our model prefers to using a 3-dimensional convolutional layer to deal with the ``text video". The size of the convolutional 
kernel is $w\times h \times n$, where $n$ is the number of words that the kernel covers at a time. 
Hence, the 3-dimensional kernel acts as an $n$-gram detector.
The single convolution with multiple kernels produces a new feature map as shown in Figure \ref{fig:model}. After the operation of the single convolutional layer, we apply the max-over-time pooling \citep{collobert2011natural} to carry out down-sampling.  The max-over-time pooling operation in our model is different from the traditional ones that are popular in the field of computer vision.
We conduct a 1-dimensional max-pooling procedure along the time axis for each channel.
The 2-dimensional feature map is combined using the max-over-time pooling procedure 
followed by a nonlinear function activation (e.g., the ReLu function).
Finally, we flatten the feature map after the max pooling, and the 
FC layers accept the flattened vectors as inputs to make the final classification.

Table \ref{tb:com-preprocess} compares the word's visual representation with other existing word embedding schemes in terms of data preprocessing steps. From the summarization, it is clear that the char-CNN, and our method require much less preprocessing steps than the one-hot vector and distributed representation. The last two rows in Table \ref{tb:com-preprocess} summarize the sparsity and dimension of word vectors for each method.

\subsection{Network Implementation}
The network architecture can be described as follows:
\begin{enumerate}
    \item \textbf{Conv3d layer}: \textit{kernel size = (20, 131, 3), stride = (1, 1, 1), number of kernels = 50, padding = 0};
    \item \textbf{MaxPool1d layer} (the max-over-time pooling):  \textit{kernel size = 3, stride = 3, dilation = 3, padding = 0 };
    \item \textbf{FC  layer 1}: \textit{input = 1250, output = 512};
    \item \textbf{FC   layer 2}: \textit{input = 512, output = 100};
    \item \textbf{FC   layer 3}: \textit{input = 100, output = number of classes}.
\end{enumerate}
The specification \textit{stride = 3} for the \textbf{MaxPool1d} 
results in no overlaps in max-over-time pooling.

\subsection{Model Interpretation}
The proposed model has a concise structure with one convolutional layer, 
one max-pooling layer, and three subsequent FC layers.
In image processing, a convolution between a kernel and local pixels 
(usually covering a $3\times 3$ dimensional area) of an image can blur, 
sharpen, emboss the image or detect edges of this image. This process 
is often applied to the neighbourhood of the local area repeatedly.

Different from the traditional way that the kernel focuses 
on local pixels of an image, we propose to compute the weighted average 
between the convolutional kernel and the whole word image as shown in Figure \ref{fig:model}. It suggests that the size of the convolution kernel should be the same as the size of the images. Because of the video-like representation of the text data, when the 3-dimensional kernel slides over the text tensor, it computes the convolutional weighted average for several word images at one time. We prefer to such kind of global convolution rather than the local convolution for the following reasons: First,
the information of text images is centralized; second, this design makes it  very convenient to interpret the convolutional operation as an $n$-gram detector.

As shown in Figure \ref{fig:model}, every time when the convolutional kernel slides over
the text, it operates on two neighbouring word images, and in this case it is a 2-gram detector. During the training, we input a sentence or an article which contains $l$ words, and the first layer of the proposed model would output $n$-gram feature vectors $\mathbf{u} \in \mathbb{R}^{l-1}$ sequentially. 
Some of the high-frequency $n$-grams of the corpus  
can be repeatedly detected by the 3-dimensional kernel. Therefore, the values of the corresponding components in the feature vector $\mathbf{u}$ for those high-frequency $n$-grams are larger than others. 
In contrast, the components in the feature vector $\mathbf{u}$ that corresponds to
the low-frequency word pairs would be small.
By applying $k$ different kernels, we can obtain a feature map $\mathbf{U}\in \mathbb{R}^{k\times (l-1)}$ as the output of the first layer as shown in Figure \ref{fig:model}. The columns of $\mathbf{U}$ are the $n$-gram features and the rows correspond to the $k$ channels.

For the testing, by inputting a test sentence, a corresponding feature map $\mathbf{U}$ is produced by the first layer of the trained model. As stated earlier, a larger value of $\mathbf{U}_{i,j}$ indicates that the $j$-th 2-gram of this test sentence is more frequently detected by the $i$-th filter, where $i=1,\cdots,k$ is the index of kernel, $j=1,\cdots,(l-1)$ is the index of 2-gram phrase.

\begin{table*}[!ht]
  \caption{Descriptions of the training, validation, and testing samples split of the five datasets.}
  \label{tb:train-test-split}
  \centering
  \setlength{\tabcolsep}{.5mm}
  \begin{tabular}{llllcccc}
    \toprule
    % \multicolumn{}{c}{Part}                   \\
    % \cmidrule(r){1-2}
 Datasets   &  Training   & Validation & Testing & Classes & Ave length & Content\\
    \midrule
    AG & 96,000 & 24,000 & 7,600 & 4 & 43 & Title$+$description fields    \\
    DBPedia   &  448,000  & 112,000 & 70,000 &14 & 52 & Title$+$abstract of article    \\
    Yelp  & 520,000 & 130,000 & 50,000 &5 & 149 & Food reviews\\
    Yahoo  & 1,120,000 & 280,000 & 60,000 &10 & 105& Title$+$question$+$best answer\\
    Amazon  & 2,880,000 & 720,000 & 400,000 &2 & 86& Title$+$content\\
    \bottomrule
  \end{tabular}

 {\footnotesize  ``Classes'' represents the number of classes, and ``Ave length''
 refers to the average number of words in the content.}
\end{table*}

\section{Evaluation}
\subsection{Baselines}
For comparisons, we consider four baseline methods as follows:
\begin{itemize}
    \item The character-level convolutional neural networks (char-CNN) \citep{zhang2015character}.
    \item CNN for text classification on top of the one-hot word vectors denoted as CNN one-hot;
    \item CNN for text classification on top of the distributed word vectors obtained via word2vec \citep{kim2014convolutional}  denoted as CNN wor2vec;
    \item  FastText \citep{joulin2017bag}.
\end{itemize}
We experiment with two variants of the proposed model:
\begin{itemize}
    \item Our model with the max-over-time pooling, as shown in Figure \ref{fig:model}; \item Our model by substituting the 1-dimensional max-over-time pooling with a 2-dimensional max pooling. The kernel size is $3 \times 3$.
\end{itemize}

\subsection{Datasets}
Five datasets used in our experiments are described as follows:
\begin{itemize}
    \item The \textbf{AG's news} corpus is a collection of more than 1 million news articles \footnote{https://www.di.unipi.it/\~gulli/AG\_corpus\_of\_news\_articles.html}. In \citep{zhang2015character}, four largest classes, namely 
        \textit{World, Sports, Business, and Science/Technology}, are selected from this corpus. Each sample is constructed by joining the title and description fields.

% The number of samples for training/validation/testing are as shown in Table \ref{tb:train-test-split}.

    \item The \textbf{DBPedia ontology} dataset 
    \footnote{https://wiki.dbpedia.org/services-resources/dbpedia-data-set-2014}. The DBpedia dataset 
    uses a large multi-domain ontology which has been derived from Wikipedia \citep{lehmann2015dbpedia}. 
    The DBpedia ontology dataset is constructed by picking 14 non-overlapping classes: 
    \textit{Company, Educational Institution, Artist, Athlete, Office Holder, Mean Of Transportation, 
    Building, Natural Place, Village, Animal, Plant, Album, Film, Written Work}. For each of these 
    14 ontology classes, the fields of samples we used are the joint of the 
    title and abstract of each Wikipedia article.
    % Table \ref{tb:train-test-split} shows the number of samples for training/validation/testing.

    \item \textbf{Yelp reviews}. The Yelp reviews dataset is obtained from the Yelp Dataset Challenge in 2015. Each review of this dataset has one user's review score ranging from 1 star to 5 stars.
    Predicting the number of users' review stars corresponds to a 5-class classification task.
    % The number of samples for training/validation/testing are as shown in Table \ref{tb:train-test-split}.

    \item \textbf{Yahoo} Answers corpus. This corpus is extracted from the \textit{Yahoo! Answers Comprehensive Questions and Answers version 1.0}. We follow \citep{zhang2015character} to construct a topic classification dataset from this corpus by selecting 10 main categories: \textit{Society, Culture Science, Mathematics  Education, Reference Computers, Internet Sports Business, Finance, Entertainment, Music Family, Relationships Politics, Government
    }.
    % The training/validation/testing splitting is as shown in Table \ref{tb:train-test-split}.

    \item The \textbf{Amazon reviews} dataset. We obtain the Amazon review dataset from the Stanford Network Analysis Project (SNAP), which spans over 18 years with 34,686,770 reviews from 6,643,669 users on 2,441,053 products \citep{mcauley2013hidden}. Different from the Yelp review dataset, we predict the binary sentiment label for each review in the Amazon dataset. The sentiment classes of reviews with 1 or 2 stars are labelled as negative, and those with 3 or 4 stars are labelled as positive.
    The samples split for training, validation, and 
    testing for all the five datasets are shown in Table \ref{tb:train-test-split}.
\end{itemize}

\begin{figure}[htp!]
    \centering
    \includegraphics[width=8cm]{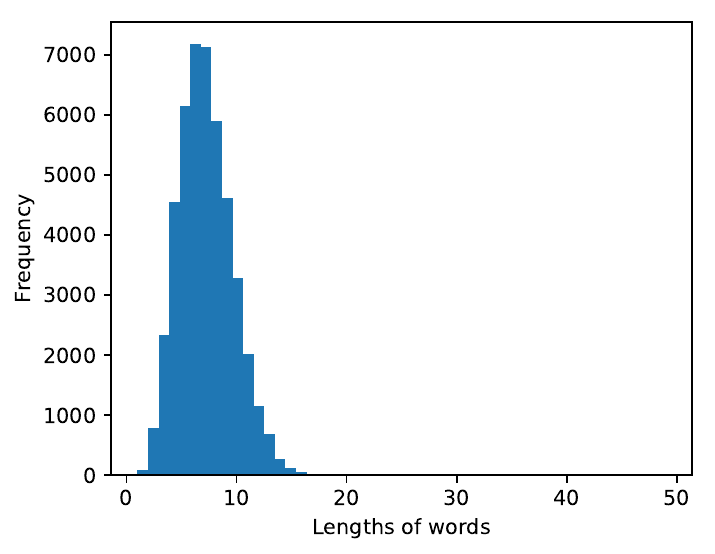}
    \caption{The histogram of the lengths of English words in the training corpus.}
    \label{fig:word_len}
\end{figure}

\subsection{Training setting}
For the implementation of our model, we need to render words into word images with their shapes. Unfortunately, the lengths of English words vary dramatically and sometimes can be very large as shown in Figure \ref{fig:word_len}. In particular, some words in the web context can have over 50 characters. For word image rendering, we adapt the size of images with the longest word. If the maximum length is too large, it can increase the blank space of word images, which causes redundancy for short words.

For English, most of the lengths of the words are less than 17 as shown in Figure \ref{fig:word_len}.
To balance the performance and redundancy, words with lengths greater than 17 are removed from the corpus. That is, we set the maximum length of words in our corpus as 17. With this threshold, we render each word in our corpus into $20\times 131$ word images. Here, the 131 pixels is the minimum width that can load 17 English characters with a font size of 20. The font we used for English characters is ``New Times".

We use the Adam \citep{kingma2014adam} algorithm as the network optimizer with 
the learning rate equal to 0.0001. The dropout rate of the FC network is 0.5.
% The numbers of samples for training, validation, and testing are as shown in Table \ref{tb:train-test-split}.

\begin{figure}[htp!]
    \centering
    \includegraphics[width=8cm]{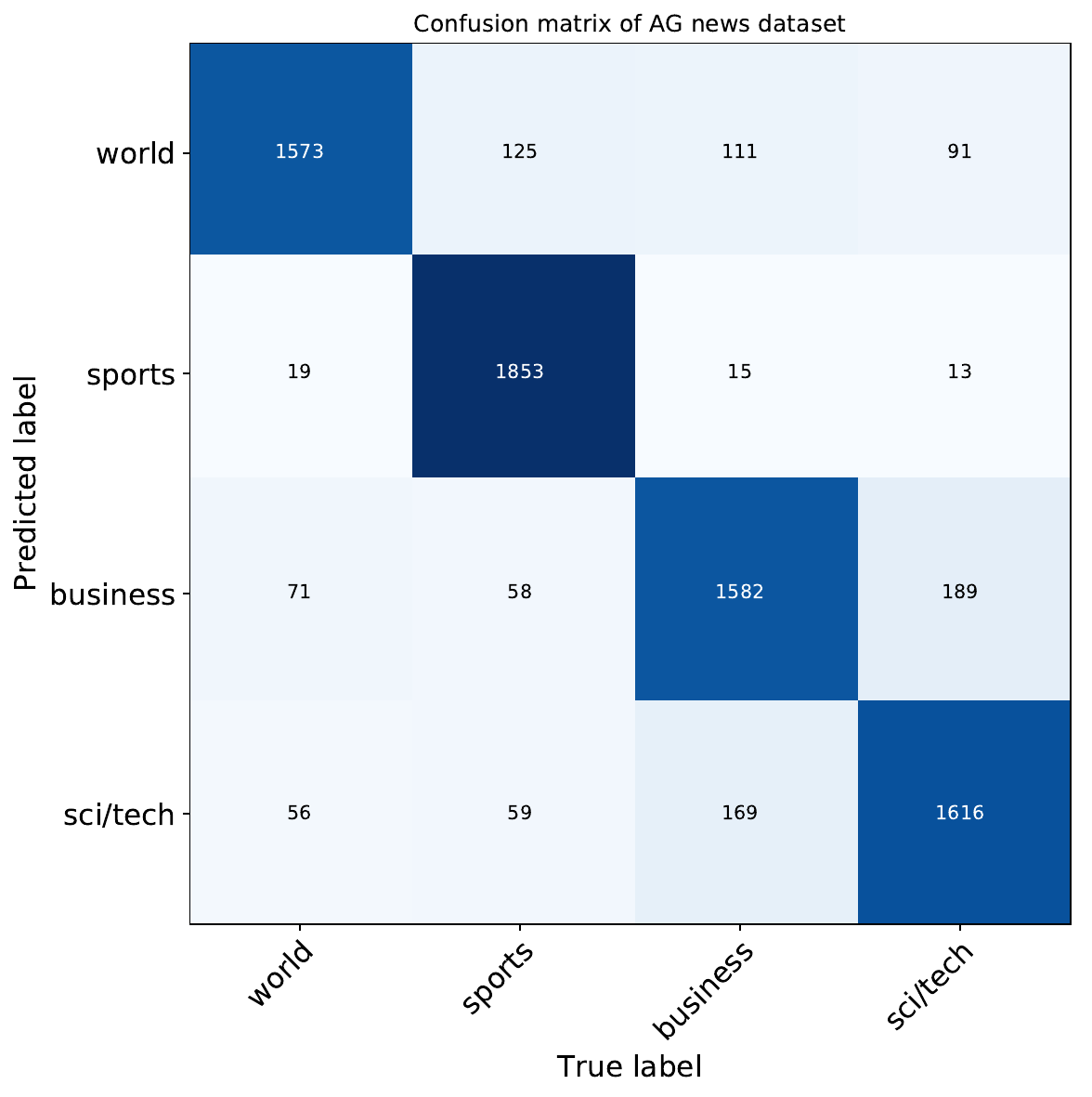}
    \caption{The confusion matrix of text classification on the AG's new dataset. From the summarization of the classification results, it can be seen that the samples belonging to the class  ``\textit{Sports}" achieve the highest correct prediction, which is consistent with the 2-gram phrase cloud in Figure \ref{fig:wordCloud} (b). }
    \label{fig:cm}
\end{figure}

\subsection{Prediction for Topics and Sentiments}
The first experiment concerns the overall performance of text classification on both the document's topics and sentiments. Table \ref{tb:acc} shows the testing accuracy for all the five 
listed datasets under all the models. The first four datasets have multiple categories, 
and the last dataset ``Amazon" has binary labels of review's polarity.

Clearly, the proposed methods achieve superior performances on the text classification compared with the existing ones for all five datasets. It is worth emphasizing that both our models (with or without max-over-time pooling) accept the text images as inputs without any preprocessing steps required by
other approaches such as removing misspellings, low-frequency words, stop words, stem and lemmatization, maintaining a vocabulary for words or characters, etc. Our methods also do not need to 
pre-train word vectors as the word2vec based methods.
The results of the two variants of our model demonstrate 
that the max-over-time pooling is an efficient and necessary 
operation for text feature extraction. Furthermore, compared with the existing methods, 
the proposed model is much more interpretable, as will be detailed in the next section.

\begin{table}[!ht]
  \caption{Comparisons of accuracy between our
  new methods (with or without
  max-over-time pooling) and four baseline methods.}
  \label{tb:acc}
  \centering
  \scalebox{0.9}{
  \begin{tabular}{lccccc}
    \toprule
    % \multicolumn{}{c}{Part}                   \\
    % \cmidrule(r){1-2}
   Method   & AG & DBPedia & Yelp & Yahoo & Amazon\\
    \midrule
    New & 0.87 & 0.98 & 0.77 & 0.57 & 0.94 \\
    New w/o max  & 0.83 & 0.93 & 0.75 & 0.54 & 0.91    \\
    char-CNN     & 0.86 & 0.84 & 0.77 & 0.56 & 0.94 \\
    CNN one-hot  & 0.86 & 0.97 & 0.69 & 0.54 & 0.89 \\
    CNN word2vec &0.85  & 0.97 & 0.68 & 0.54 & 0.91 \\
    FastText     &0.84  & 0.94 & 0.73 & 0.54 & 0.88 \\
    \bottomrule
  \end{tabular}
  }
\end{table}

\begin{figure*}[!ht]
        \centering
        \begin{subfigure}[b]{0.475\textwidth}
            \centering
         \includegraphics[width=1\textwidth]{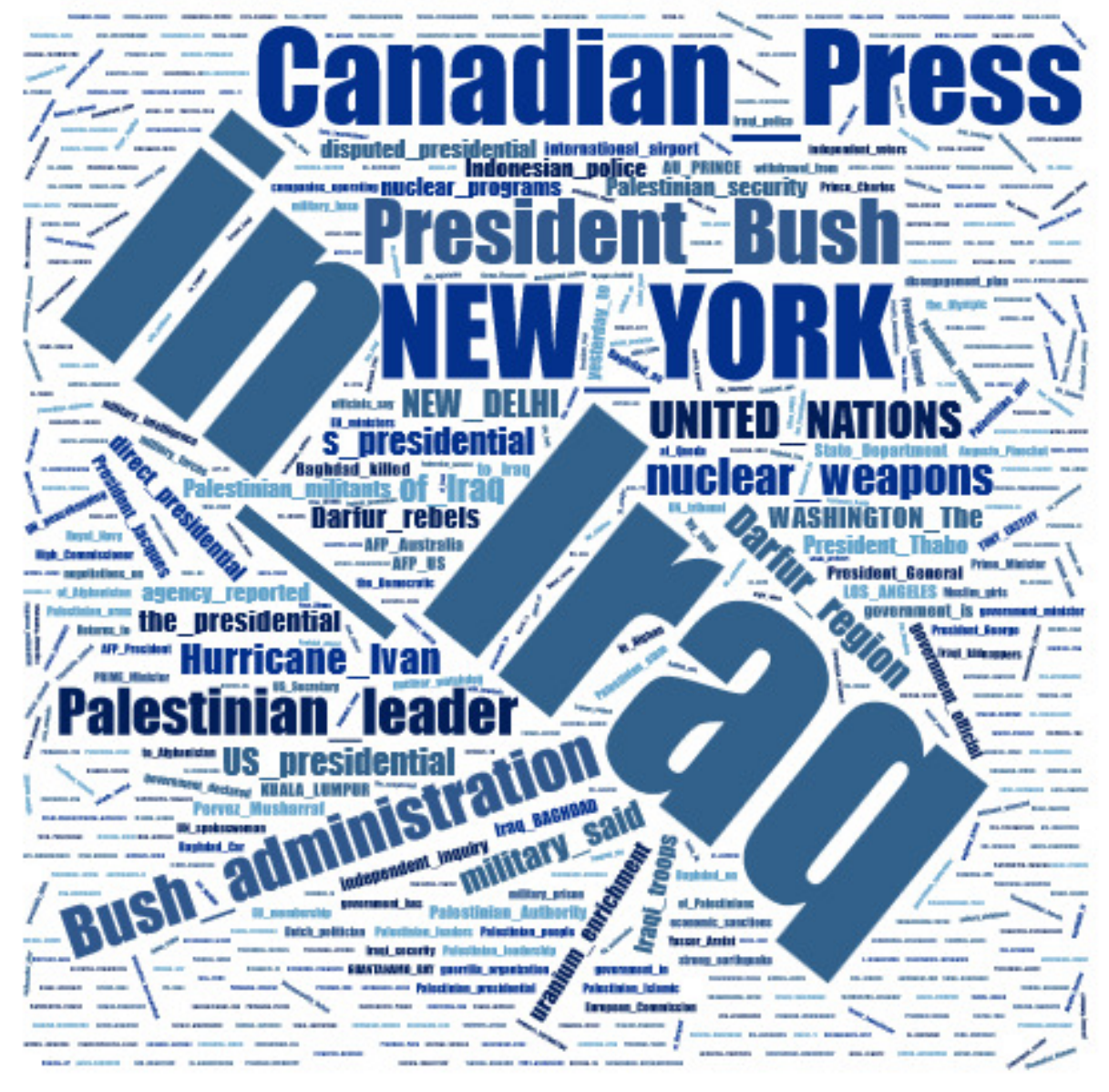}
            \caption[Network2]%
            {{\small Category: World}}
            \label{fig:fig3a}
        \end{subfigure}
        \hfill
        \begin{subfigure}[b]{0.475\textwidth}
            \centering
            \includegraphics[width=1\textwidth]{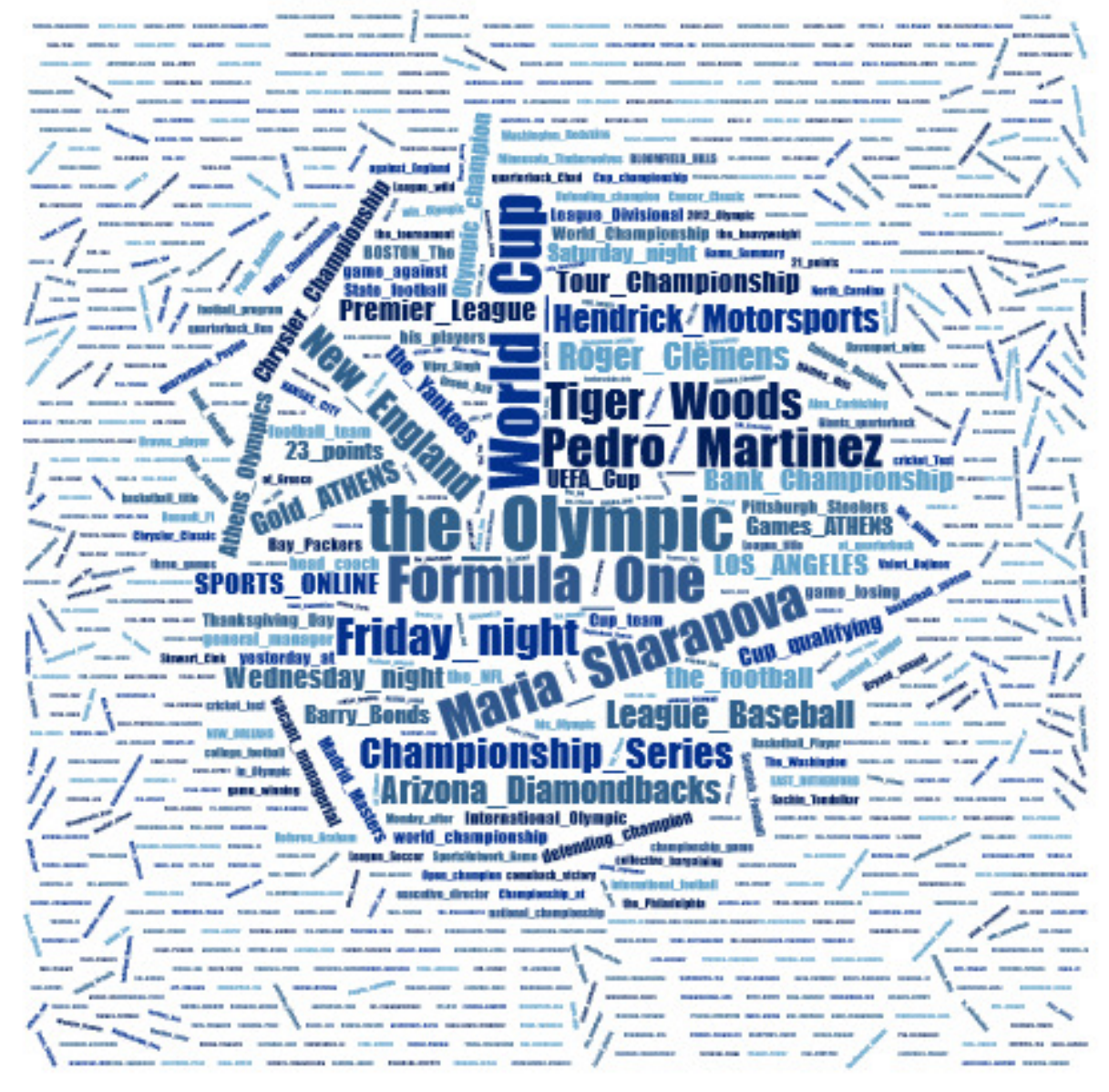}
            \caption[]%
            {{\small Category: Sports}}
            \label{fig:fig3b}
        \end{subfigure}
        \vskip\baselineskip
        \begin{subfigure}[b]{0.475\textwidth}
            \centering
            \includegraphics[width=1\textwidth]{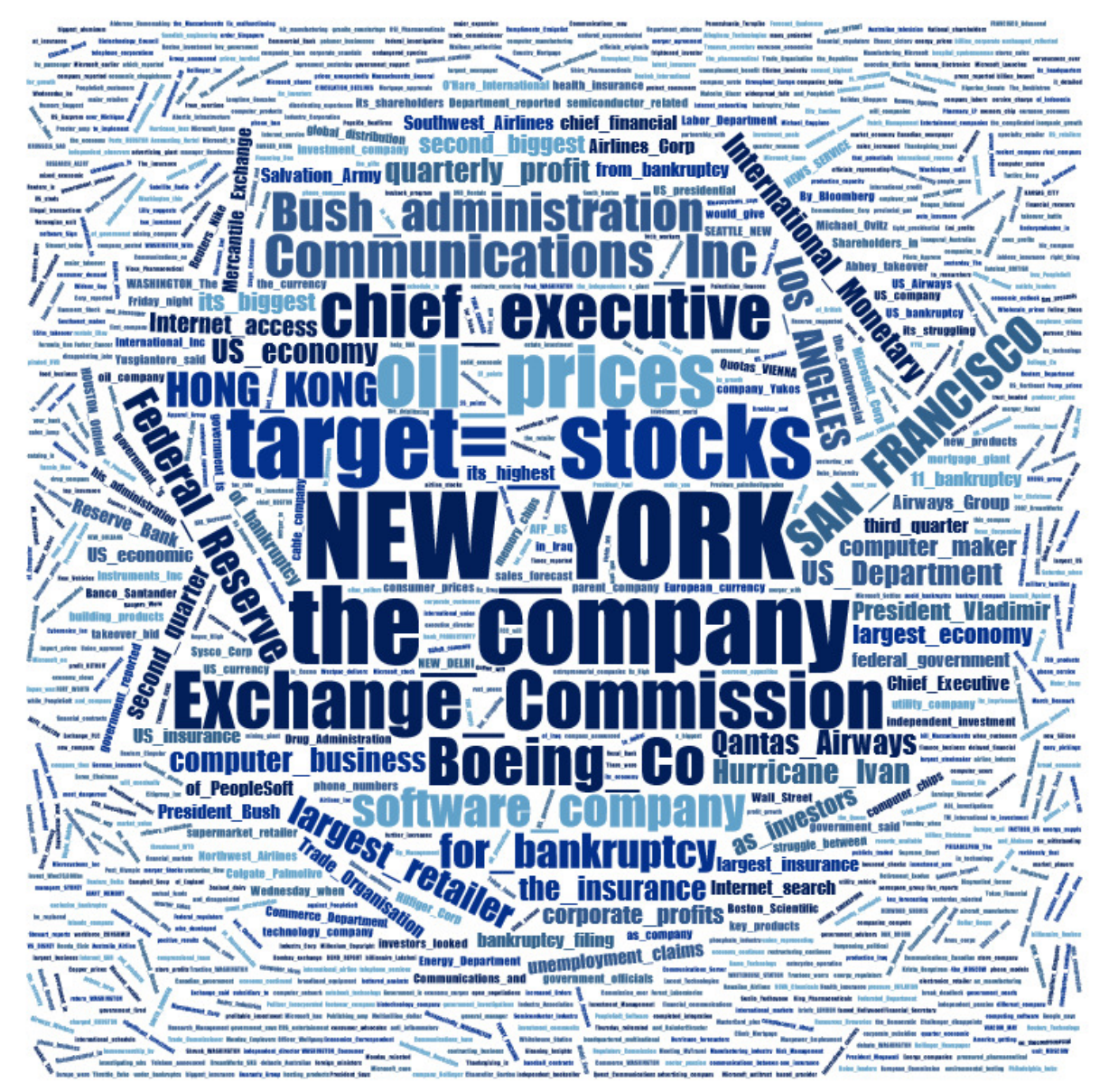}
            \caption[]%
            {{\small Category: Business}}
            \label{fig:fig3c}
        \end{subfigure}
        \quad
        \begin{subfigure}[b]{0.475\textwidth}
            \centering
            \includegraphics[width=1\textwidth]{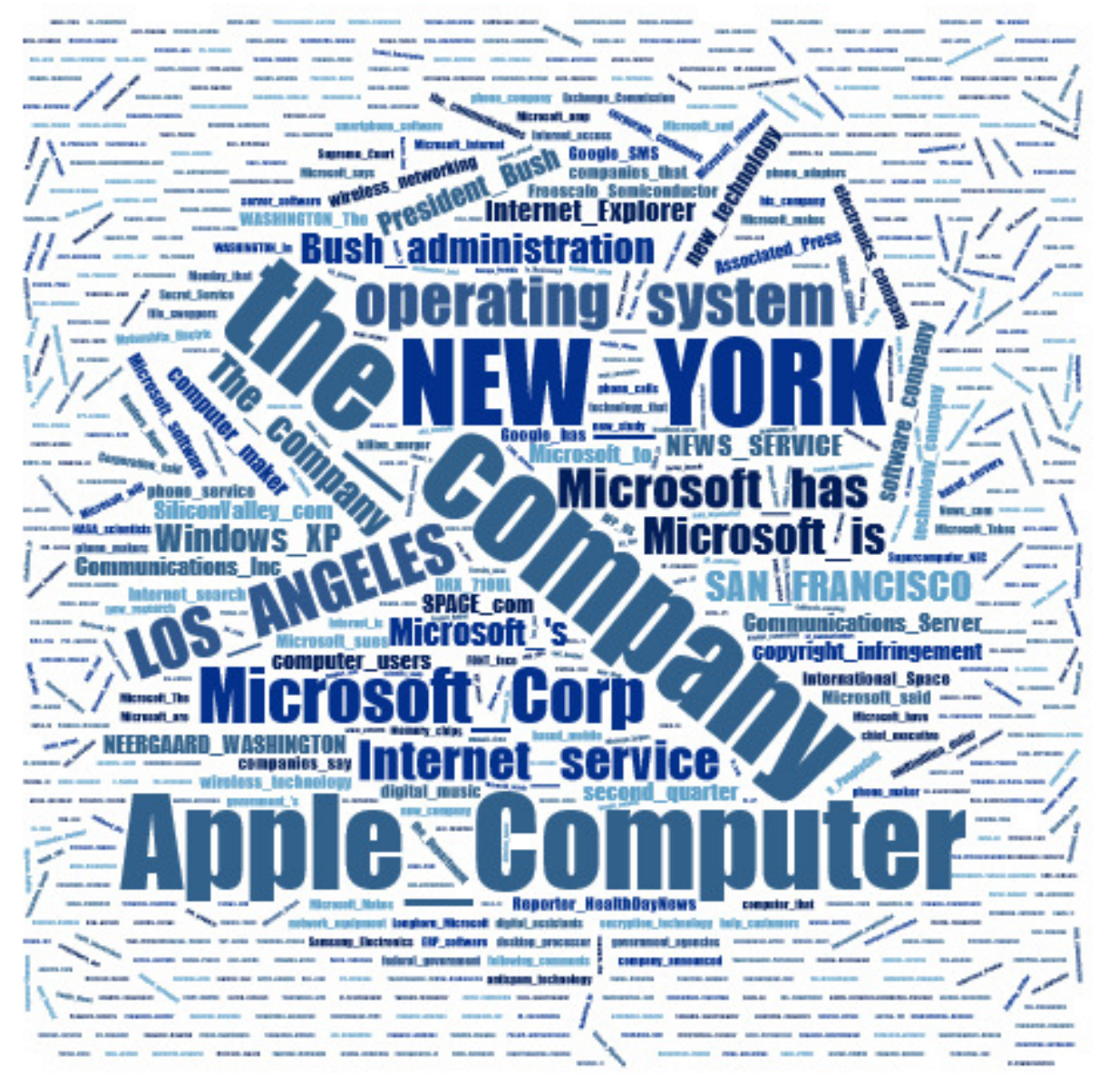}
            \caption[]%
            {{\small Category: Science and Technology}}
            \label{fig:fig3d}
        \end{subfigure}
        \caption[ The average and standard deviation of critical parameters ]
        {\small The two-word phrase clouds for four categories of the AG's news dataset. These text cloud pictures display the ``most-weighted" word pairs in the testing corpus by our 2-gram detectors. There are four categories: World, Sports, Business, and  Science \&  Technology. The maximum confusion of (a) and (d) is ``NEW\_YORK"; the maximum confusion of (c) and (d) is ``the\_company". We add an underscore ``\_" to combine two words in a 2-gram phrase. }
        \label{fig:wordCloud}
    \end{figure*}

\subsection{Visualization of the $n$-gram detectors}
The 3-dimensional convolutional kernel acts as an $n$-gram detector in our model. As shown in Figure \ref{fig:model}, the conventional kernel operates on two frames (i.e., two words) at a time, 
which thus corresponds to a 2-gram detector. For a sentence $S$ of length $l$, we can 
generate a feature vector $\mathbf{u} \in \mathbb{R}^{l-1}$, which is a continuous $2$-gram feature of $S$.
By applying $k$ different kernels, we can obtain a feature map $\mathbf{U}\in \mathbb{R}^{k\times (l-1)}$  after the 3-dimensional convolution.

During the testing, we input a test sentence, 
and a corresponding feature map $\mathbf{U}$ is the output. A larger value of $\mathbf{U}_{i,j}$ indicates that the $j$-th 2-gram of the input sentence is more significant for the classification.
By identifying the maximum element $\mathbf{\widetilde{U}}_{i,j}$ of $\mathbf{U}$, we can easily identify the most-weight word pair (the $j$-th word and $(j+1)$-th word) in this sentence $S$ reversely, where $j+1\leq l$, $l$ is the number of words in this sentence.

We visualize the weighted $n$-grams according to the first layer of the network trained on the task of classifying the AG's news dataset. It has four classes, ``\textit{World}", ``\textit{Sports}", ``\textit{Business}", and ``\textit{Science \& Technology}". There are 7600 test samples for all categories and each class has 1900 samples. Because all the testing 2-grams have been weighted by the feature map $\mathbf{U}$ for each class, we can visualize them with two-words phrase (2-grams) clouds separately as shown in Figure \ref{fig:wordCloud} (a)--(d). A larger font size in the word cloud pictures indicates a higher frequency that this two-words phrase has been detected by the 3-dimensional $2$-gram detectors. The two words of these 2-grams are joined by an underscore ``\_" for the convenience of visualization.

According to the results of Figure \ref{fig:wordCloud} (a), we observe that the $2$-gram ``\textbf{in Iraq}" is the most frequent phrase that has been detected for the category ``\textit{World}" news. Some other phrases such as ``\textbf{Canadian Press}", ``\textbf{NEW YORK}", ``\textbf{President Bush}", ``\textbf{UNITED NATIONS}" and so on that are associated with the category ``\textit{World}" news, have also been highlighted by our 2-gram detectors. In Figure \ref{fig:wordCloud} (b)--(d), we can also see that the 2-grams ``\textbf{World Cup}", ``\textbf{the Olympic}", ``\textbf{Formula One}" have been detected for the category ``\textit{Sport}"; ``\textbf{the company}", ``\textbf{target= stocks}", ``\textbf{Oil prices}" and so on have been detected for the category ``\textit{Business}"; and ``\textbf{the company}", ``\textbf{Apple Company}" and so on have been detected for the category ``\textit{Science \& Technology}".

By comparing Figure \ref{fig:wordCloud} (a) with Figure \ref{fig:wordCloud} (d), we observe that ``\textbf{NEW YORK}" is the intersection of the high-frequency phrase between the most-weighted 2-grams for categories ``\textit{World}" and ``\textit{Science \& Technology}". It suggests that there might be ambiguity for our 2-gram detector when classifying the phrase 
``\textbf{NEW YORK}" in the categories of ``\textit{World}" news or ``\textit{Science \& Technology}" news. 
%which may cause misclassifications. 
The same situation arises when categorizing ``\textit{Business}" and ``\textit{Science \& Technology}". By comparing Figure \ref{fig:wordCloud} (c) with Figure \ref{fig:wordCloud} (d), we find that the most-highlighted 2-grams for ``\textit{Business}" and ``\textit{Science \& Technology}" also have an intersection of ``\textbf{the company}", which clearly belong to both categories. In contrast, the highlighted 2-grams in Figure \ref{fig:wordCloud} (b) has no intersection with other three categories, which makes the category ``\textit{Sports}" most 
%easily to be classified 
distinctive from the others. The confusion matrix of the four-class classification 
in Figure \ref{fig:cm} supports this argument.

\section{Conclusion}
We propose a novel framework to understand the text data by converting 
English sentences or articles into a video-like 3-dimensional tensors,
which can be viewed as ``video text''. Each frame or each slice of the tensor 
is a word image that is rendered as the word's shape. This transformation makes it  convenient to implement an $n$-gram model based on the convolutional neural networks. We achieve this goal by imposing a 3-dimensional convolutional kernel on text tensors. The first two dimensions of the kernel size are 
the same as the size of the word image and the last dimension of the kernel size is $n$. That is,  the 3-dimensional kernel covers $n$ words and outputs a scalar each time. A subsequent 1-dimensional max-over-time pooling is applied to this feature map, and then three FC layers are 
implemented with a final goal for text classification.
Experiments of text classification on both topic and sentiment analysis 
illustrate surprisingly excellent results of the proposed model.
Our model can be easily applied to other languages as well as 
%In the future, we plan to extend our model 
other NLP tasks such as the machine translation.

\bibliographystyle{named}
\bibliography{sample}

\end{document}